\documentclass[letterpaper, 10 pt, journal, twoside]{IEEEtran}
\IEEEoverridecommandlockouts

\usepackage{cite}

\usepackage{url}

\usepackage{amsmath,amssymb,amsfonts}
\usepackage{algorithmic}
\usepackage{graphicx}
\graphicspath{ {./images/} }

\usepackage{tabularx, booktabs}
\usepackage{multirow}
\usepackage[footnotesize]{caption}

\usepackage{dirtytalk}

\usepackage{algorithm}
\usepackage{footnote}

\usepackage[caption=false, font=footnotesize]{subfig}

\usepackage{etoolbox}

\usepackage{gensymb}
\usepackage{textcomp}
\usepackage{xcolor}

\def\BibTeX{{\rm B\kern-.05em{\sc i\kern-.025em b}\kern-.08em
    T\kern-.1667em\lower.7ex\hbox{E}\kern-.125emX}}
\begin{document}

\pagenumbering{gobble} 

\markboth{IEEE Robotics and Automation Letters. Preprint Version. Accepted June, 2022}{Oh \MakeLowercase{\textit{et al.}}: Monocular UAV Localisation with Deep Learning and Uncertainty Propagation} 

\title{Monocular UAV Localisation with Deep Learning and Uncertainty Propagation}

\author{Xueyan Oh, Ryan Lim, Leonard Loh, Chee How Tan, Shaohui Foong, and U-Xuan Tan

\thanks{Manuscript received: February, 24, 2022; Revised June, 15, 2022; Accepted June, 15, 2022.}
\thanks{This paper was recommended for publication by Editor Sven Behnke upon evaluation of the Associate Editor and Reviewers’ comments.
This research was supported by the National Robotics R\&D Programme Office and Agency for Science, Technology and Research (A*STAR) under its National Robotics Programme - Robotics Enabling Capabilities and Technologies (W1925d0056).} 
\thanks{The authors are with the Engineering Product Development Pillar, Singapore University of Technology and Design, Singapore
        {\tt\footnotesize xueyan\_oh@mymail.sutd.edu.sg}}%
\thanks{Digital Object Identifier (DOI): see top of this page.}
}

\makeatletter
\patchcmd{\@maketitle}
  {\addvspace{0.5\baselineskip}\egroup}
  {\addvspace{-2\baselineskip}\egroup}
  {}
  {}
\makeatother

\maketitle

\begin{abstract}
In this paper, we propose a ground-based monocular UAV localisation system that detects and localises an LED marker attached to the underside of a UAV. Our system removes the need for extensive infrastructure and calibration unlike existing technologies such as UWB, radio frequency and multi-camera systems often used for localisation in GPS-denied environment. To improve deployablity for real-world applications without the need to collect extensive real dataset, we train a CNN on synthetic binary images as opposed to using real images in existing monocular UAV localisation methods, and factor in the camera’s zoom to allow tracking of UAVs flying at further distances. We propose NoisyCutout algorithm for augmenting synthetic binary images to simulate binary images processed from real images and show that it improves localisation accuracy as compared to using existing salt-and-pepper and Cutout augmentation methods. We also leverage uncertainty propagation to modify the CNN’s loss function and show that this also improves localisation accuracy. Real-world experiments are conducted to evaluate our methods and we achieve an overall 3D RMSE of approximately 0.41m.
\end{abstract}

\begin{IEEEkeywords}
Localisation, Aerial Systems: Applications, Visual Tracking
\end{IEEEkeywords}

\vspace{-3mm}

\section{Introduction}
\IEEEPARstart{T}{he} use of UAV for real-world tasks such as maintenance, assessing damage, site inspection and safety inspection in civil applications is becoming increasingly popular recently due to advantages such as in cost, efficiency, safety, and mobility \cite{R2,R3,R4}. However, accurate and reliable localisation for autonomous UAV tasks in GPS-denied environment such as indoors or outdoors near structures is still challenging \cite{R2,R5,R6,R7,R8}. Technologies such as those based on Radio Frequency (RF) signals require infrastructure and can be unreliable in such environments due to reflection, interference, or degradation of RF signals by walls and other obstacles \cite{R2, R7}. Although ultra-wideband (UWB) technology is commonly used in localisation systems and can provide up to centimetre-level accuracy, it requires setting up and calibration of multiple anchors and have limited working range as accuracy substantially decreases when the tracked object is further (just several metres) from the anchor system’s centre of mass \cite{R9}.

\begin{figure}[t]
\centerline{\includegraphics[width=0.8\columnwidth]{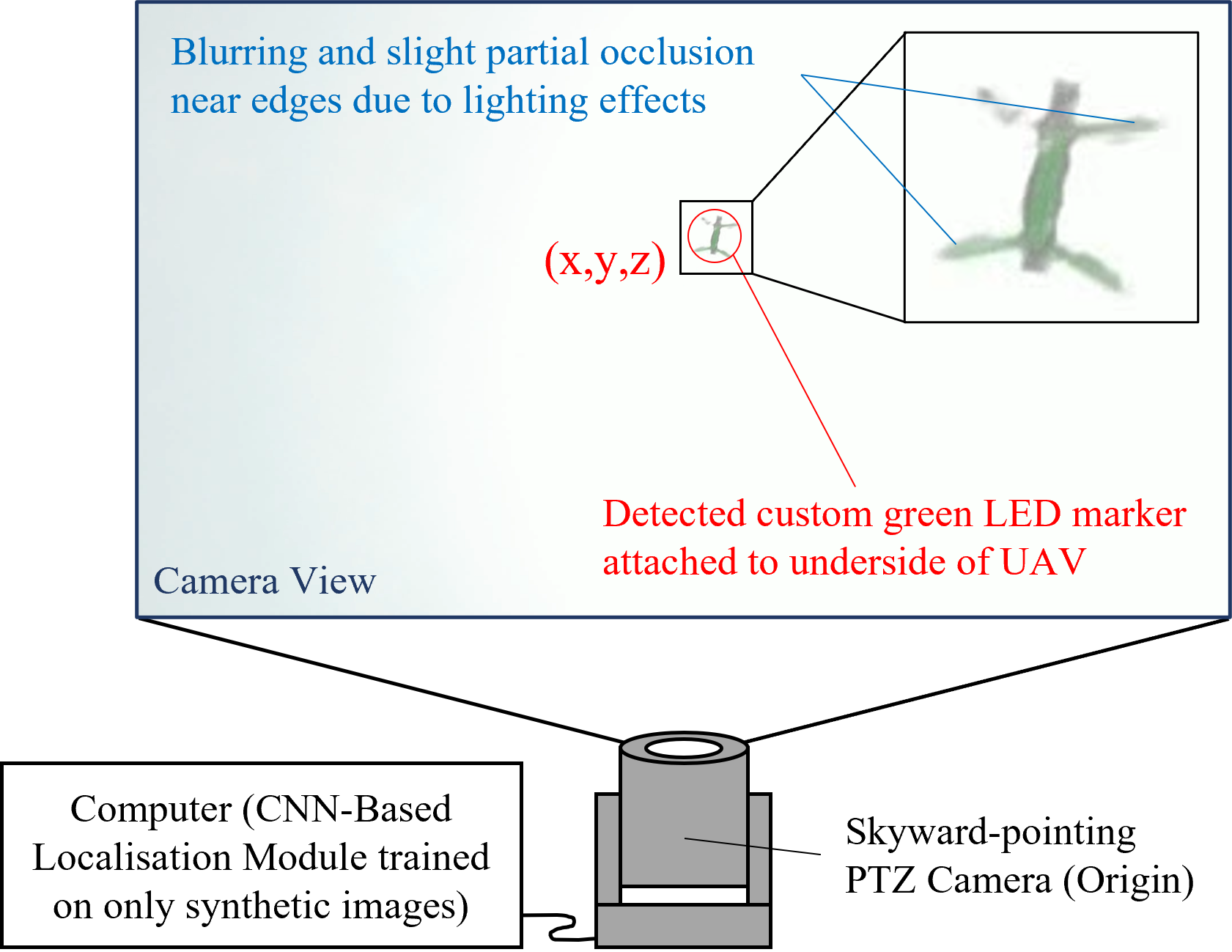}}
\vspace{-1mm}
\caption{Setup of proposed ground-based UAV localisation system.}
\label{fig:front}
\vspace{-7mm}
\end{figure}

Vision-based approaches are becoming more popular for UAV localisation in GPS-denied environments as they are lower in weight and cost as well as being versatile to scenes and can provide high accuracy \cite{R7, R10}. We refer to off-board relative visual localisation of a UAV as the continuous estimation of a UAV’s pose with respect to the camera and does not suffer from positioning drift unlike methods such as optical flow. This contrasts with on-board visual localisation where the processing of images obtained from a camera mounted on a UAV are subjected to the computational limitations of the on-board CPU \cite{R2,R5,R10}. Off-loading resources used for localisation also frees up resources for other payloads and on-board computational tasks.

Other ground-based, off-board visual approaches use visual markers and multiple cameras. Commercial motion capture systems such as OptiTrack \cite{R19} and Vicon \cite{R20} use off-board infrared cameras to track reflective markers with high accuracy but suffer from limitations such as multi-camera placement and calibration, interference of sunlight, and installation of expensive cameras \cite{R10}. Other works \cite{R5,R22} explore monocular relative localisation by attaching markers to robots and obtaining the markers' position within the camera image for point correspondence and pose estimation but have only been tested in simulation or controlled environments. Fiducial markers such as AprilTag \cite{R23} and ArUco \cite{R24} are commonly used for marker-based visual pose estimation tasks and have been used for UAV localisation \cite{R25,R26}. However, they require a large unobstructed area on the target object and good lighting for detection while attaching large fiducial markers to the underside of UAVs also interferes with their aerodynamics \cite{R28}. Real-world UAV applications require high localisation accuracy to prevent collisions when flying near structures and often operate at heights beyond 10 m to reach structures such as bridges and power lines. Visual localisation approaches are also easily affected by blurring and occlusions-like effects due to lighting conditions \cite{survey,looming}.

Recently, deep learning has been used to improve off-board monocular UAV localisation due to its robustness to unstructured environments. Jin \textit{et al.} \cite{R29} propose using relational graph networks to improve UAV keypoints detection for better 6D pose estimation but only report tests with a UAV flying up to 4m away from the camera. Unlu \textit{et al.} \cite{R30} use optical flow to detect changes in frames from a Pan-Tilt-Zoom (PTZ) camera, train a convolutional neural network (CNN) using real images to detect a UAV and obtain 3D location estimates of the UAV within 18.2\% of their ground truth. Lai and Huang \cite{R31} train a network for detection of the UAV and a separate network to estimate its distance from the camera and obtain Root-Mean-Square Errors (RMSE) of 0.65m to 2m across different scenes. Vrba and Saska \cite{R32} train a CNN to predict the position and size of detected bounding sphere of marker-less UAVs within the camera frame to estimate its distance from the camera and combine with depth-sensor data to obtain a RMSE of 3.76m. However, the accuracies obtained in these works may be insufficient for UAV tasks that require more precise maneuverability such as flying close to a structure for inspection.

Existing deep learning-based methods often train on real images of UAV. However, real-world UAV applications often use custom-built UAVs with specialised payloads, and it is expensive to obtain new real images with their labels to retrain the CNN whenever there is a change in the UAV’s appearance, making it less deployable for real-world applications. To address the above limitations, we propose to track an LED marker attached to the underside of a UAV using a ground based PTZ camera pointed upwards towards the UAV’s area of operation. We use a background subtractor and train a CNN to detect and predict the size and location of the marker within each camera frame in real-time. Our main contributions are as follows:

\begin{itemize}
\item We propose a more deployable deep learning-based monocular UAV localisation system that uses an LED array as a visual marker that can be attached to the underside of a UAV and show that we can remove the expensive process of obtaining real images by training a CNN on purely synthetic images for detection and localisation of its marker.
\end{itemize}
\begin{itemize}
\item We propose NoisyCutout algorithm to augment binary images and demonstrate that a CNN trained on synthetic binary images augmented using NoisyCutout improves predictions of the UAV’s marker size and position from binary images processed from real video frames.
\end{itemize}
\begin{itemize}
\item We propose a loss function that leverages uncertainty propagation using the relationships between errors in pixel predictions and their respective estimated Cartesian coordinates and demonstrate that it improves localisation accuracy in real-world experiments.
\end{itemize}

\section{Related Work}

Learning-based approaches for visual object pose estimation out-perform classical non-learning vision-based methods in challenging scenes such as messy backgrounds, occluded objects, large illumination changes and lack of textures \cite{R33} and recent works use them for UAV applications \cite{R32,R34,R36}. Seidaliyeva \textit{et al.}\cite{R36} use background subtraction to extract regions of interest as input into a CNN to classify whether a UAV is present. However, the CNN is trained using manually labelled real images which is very time consuming and require new images to retrain the CNN to detect custom UAVs with different appearances. Furthermore, the paper focuses on UAV detection without 3D localisation. While our work uses a similar background subtraction method to detect a UAV, we also estimate the 3D location of a UAV relative to the camera and focus more on size estimation of its attached LED marker to improve localisation accuracy.

Recently, Vrba and Saska \cite{R32} propose a micro-UAV relative visual localisation approach that trains a CNN to detect marker-less micro-UAVs and estimate their bounding box to calculate the UAVs' position and distance from the camera. However, their training dataset consists of manually labelled real images of UAVs and they obtain a high average localisation error of 2.86m. Wang \textit{et al.} \cite{R37} introduce a CNN-based monocular object detection model for 3D positioning between a ground-based camera and a UAV in GPS-denied environments and report a RMSE of 1.57m. They also estimate the UAV’s distance from the camera using its predicted size but without particularly considering edge effects due to blurring and occlusion. Our approach trains a CNN on synthetic binary images augmented to simulate these conditions. We also improve distance estimation and localisation accuracy by leveraging uncertainty propagation in the loss function used to train our CNN.

\begin{figure*}[t]
\centering
\centerline{\includegraphics[width=0.9\textwidth]{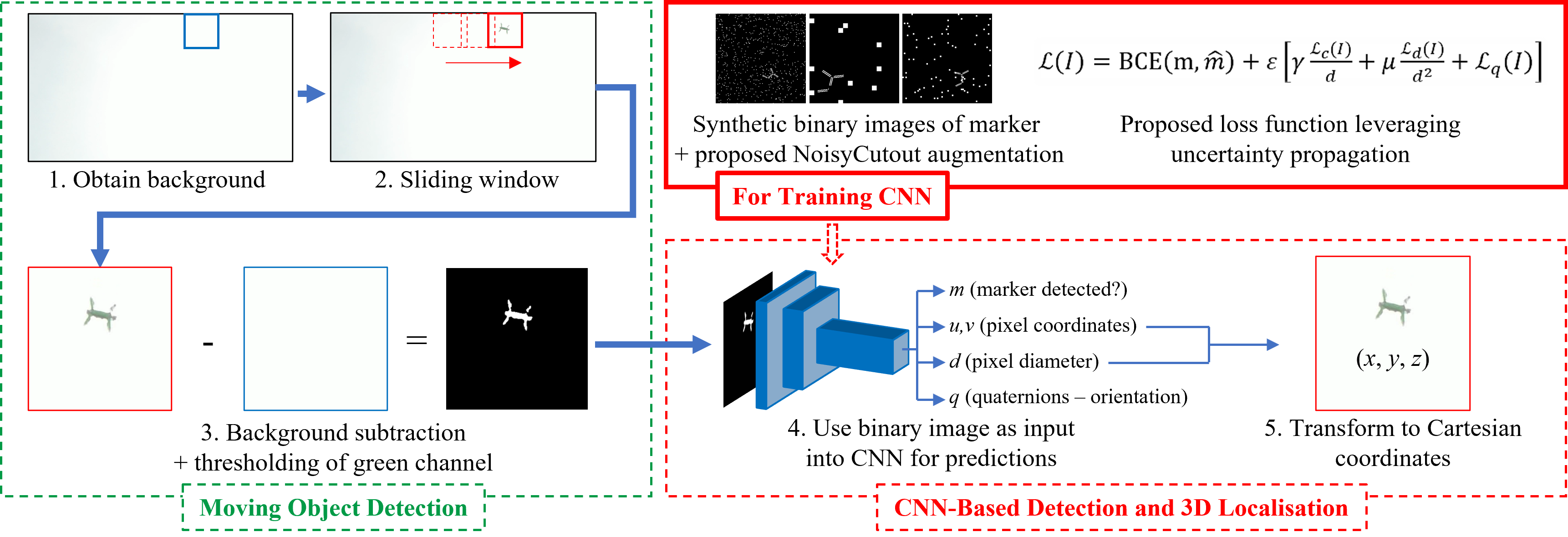}}
\vspace{-1mm}
\caption{Proposed localisation pipeline – Marker-based-UAV-Localisation-Net (MbULNet) – for localisation of a UAV with an LED marker attached to its underside. Our main technical contributions involve our proposed NoisyCutout algorithm for augmenting synthetic binary images used to train our CNN as well as a loss function that leverages uncertainty propagation (UP) to improve pixel predictions of the detected marker.}
\label{fig:pipeline}
\vspace{-6mm}
\end{figure*}

Image augmentation techniques are widely used to increase the variations in training data \cite{R38}. Recent techniques such as Cutout \cite{R1} and Random Erasing \cite{R40} disrupt images by masking sections or specific objects within images to train the CNN to learn minor object features that are usually ignored and be more robust to occlusions. On the other hand, Wright \cite{progressivesprinkles} suggests that the above methods potentially block out important features due to the large mask size and proposed Progressive Sprinkles (we refer to as PS) that uses smaller random masks such that important features are almost always present and show improved results in a classification task. Our proposed augmentation method is similar to PS but applied in a more systematic manner onto binary images as well as simulate distractors and inflationary or deflationary effects on the object observed due to lighting.

\section{Localisation System and Pipeline}

\subsection{Proposed Setup of UAV Localisation System}

Our proposed UAV localisation system consists of a custom array of LEDs that functions as a marker secured to the underside of a UAV and is observed by a ground-based camera connected to a localisation module that processes camera frames to output 3D coordinates of the UAV. Optical zoom function for the camera is optional but allows the localisation system to work at further marker-to-camera distances where the marker becomes too small in the frame without zoom. It is assumed that the camera’s vertical and horizontal field-of-view (FOV) are known at different zoom presets. The LED array can span as much of the UAV’s footprint as possible, and has no specific requirements for the spacing between each LED in the array or sources of slight partial occlusions when observed from the ground. To demonstrate our approach, we design our marker using LED strips arranged to match the frame of a UAV used in our experiments and use a ground-based PTZ camera for its ability to tilt facing skywards and its zoom. Fig.~\ref{fig:front} shows the setup of our proposed marker-based UAV localisation system.

\subsection{Localisation Pipeline}\label{locpipeline}

To detect the marker and estimate its Cartesian coordinates relative to the camera, we propose a pipeline consisting of two stages: 1) moving object detection; and 2) CNN-based classification and localisation. This pipeline is similar to the work by Seidaliyeva \textit{et al.}\cite{R36} that combines the speed and accuracy of traditional object detection methods with the robustness of neural networks. To cater our method to different camera FOV without retraining the CNN, we train our CNN to predict the marker’s pixel coordinates and diameter within the image and transform these outputs to the Cartesian coordinates of the marker relative to the camera as a function of FOV. We illustrate our proposed pipeline in Fig.~\ref{fig:pipeline} and describe the details of each stage in this section.

\textit{1) Moving Object Detection:} We use background subtraction \cite{R41} with thresholding, and combine this with a sliding window so that we can use a smaller region-of-interest (ROI) as input into a CNN in the next stage without resizing and losing resolution. Background subtraction assumes a relatively static background and involves obtaining the median of the first 50 video frames after the camera has been set up and without the marker in view. To detect a moving object, we assume that there will be no more than one marker in the camera frame, and apply a sliding window of 299 x 299 pixel with an overlap of 20 pixels that slides on every new frame. As we choose to use a green LED marker to demonstrate our approach, the difference in the green channel of each window and that of the corresponding pixels from the median background is then obtained and thresholded with a pre-determined value \begin{math}\tau\end{math}. This gives us a mask to extract green pixels from the original frame and perform another thresholding with a green pixel threshold of 100, resulting in a binary image for use as input into a CNN.

\textit{2) CNN-based Detection and 3D Localisation:} We use binary images for the CNN’s input as the LED marker captured by a camera varies largely under different real-world lighting conditions and the CNN fails to reliably detect the LED marker when trained on synthetic RGB images of our LED marker with randomised background. We find that using binary images generated from the moving object detection step as input to the CNN results in much more reliable detection of the LED marker. Our CNN predicts whether the input binary image contains the LED marker and regresses the marker’s pixel coordinates and diameter within the input image. We then transform pixel values in the image space to Cartesian space. When a marker is detected, the sliding window stops, and the predicted pixel coordinates of the marker within the full image becomes the centre of the next ROI used as input into the CNN. Conversely, the sliding window continues when no marker is detected and will eventually detect a marker if it is moving within camera view. To remove the need to train the CNN using real images, we propose to generate purely synthetic binary images of the marker with their ground truth and augment them to simulate blurring and slight partial occlusion.

\begin{figure*}[t]
\centering
\centerline{\includegraphics[width=1.0\textwidth]{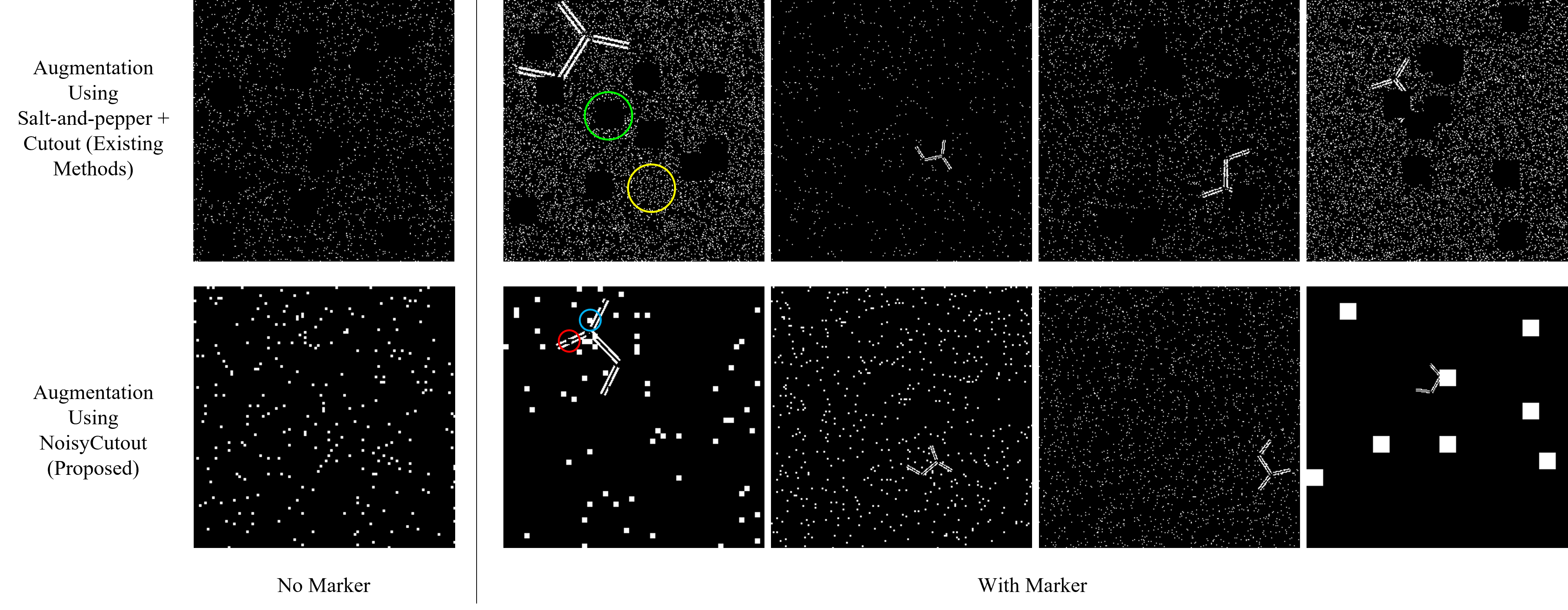}}
\vspace{-3mm}
\caption{(Top) samples of synthetic binary images augmented with salt-and-pepper noise (yellow circle) and Cutout \cite{R1} (green circle) that only applies fix-sized zero-masks. (Bottom) synthetic images augmented using our proposed NoisyCutout, applying both zero-masks (red circle) to simulate slight partial occlusions and white-masks (blue circle) of various sizes that simulate inflation of the observed marker and background distractors. These synthetic images, together with their ground truth, form the synthetic dataset we use to train our CNN for classification, localisation, and marker size estimation.}
\label{fig:syn}
\vspace{-3mm}
\end{figure*}

\section{Synthetic Data Generation and NoisyCutout}

\subsection{Synthetic Dataset Generation}

We use CoppeliaSim \cite{R42} as our simulation environment and simulate our LED marker using a group of white rectangles that matches the arrangement and dimensions of the LED strips attached to the actual UAV used for demonstrating our approach. We use a skyward-pointing (towards positive z-direction) virtual camera with a resolution of 299 x 299 pixel to capture images of the marker against a black background while varying the following:

\begin{itemize}
\item Height (along z-axis) of the marker of between +2 and +20m from the virtual camera’s origin;
\end{itemize}

\begin{itemize}
\item Pixel coordinate of the marker’s centroid while ensuring marker is always within the camera’s view;
\end{itemize}

\begin{itemize}
\item Marker’s yaw (rotate about z) between -180{\degree} and +180{\degree};
\end{itemize}

\begin{itemize}
\item Marker’s roll and pitch between -5{\degree} and +5{\degree}.
\end{itemize}

We generate 6800 synthetic images, with 5500 containing the marker and 1300 null images (with all pixel values = 0).

\setlength{\textfloatsep}{2pt}
\begin{algorithm}[t]
\caption{NoisyCutout}
\begin{algorithmic}[1]

\STATE\textbf{Input: }{Binary Image $X$ //Shape: (299 x 299 x 1)}
\STATE $s =$ random.uniform(1, 31) //$s$ = size of cutout
\STATE //Apply non-overlapping $s$ x $s$ sliding window
\FOR{window in image $X$}
    \IF{random.uniform(0,1) $< 0.05$ //cutout prob. P = 0.05}
        \IF{random.uniform(0,1) $< 0.5$}
            \STATE value of all pixels in window = 255
        \ELSE
            \STATE value of all pixels in window = 0
        \ENDIF
    \ENDIF
\ENDFOR
\STATE Apply Gaussian blur with randomly selected kernel size from [(1,1), (1,3), (3,1), (3,3)] to Image $X$
\STATE Apply thresholding to $X$: non-zero pixels = 1
\end{algorithmic}
\label{alg:noisycutout}
\end{algorithm}

\subsection{NoisyCutout}

We augment all binary images to simulate real-world effects and noise, including blurring and slight partial occlusion due to lighting conditions and features such the UAV’s own landing gear, uneven distribution of light between LEDs, and motion blur. We leverage the concepts of Cutout, salt-and-pepper noise, and Gaussian blur to form an augmentation technique we call NoisyCutout as described in Algorithm \ref{alg:noisycutout}. While Cutout and other existing variations \cite{R38} apply one or a low number of fix-sized zero-mask(s) to random position(s) within training images, NoisyCutout is different as it generates numerous n x n-pixel square masks with values of either one or zero. Similar to PS \cite{progressivesprinkles}, NoisyCutout adds small masks throughout each image but uses a non-overlapping n x n sliding window where n is randomly chosen between 1 and 31 for each image such that the largest window is smaller than the pixel size of the smallest UAV during training. This reduces the chance that the UAV may be fully masked in an image that may introduce false positives. At each window, we apply a mask at 0.05 probability, where all values are replaced with either one or zero (equal chance). We then randomly apply a slight Gaussian Blur before a final thresholding. NoisyCutout augments binary images such that the random arrangements of Cutouts with ones or zeros throughout an image form patterns that simulate effects of slight partial occlusions of the marker, slight inflation of the marker, as well as background distractors. Randomly formed clusters of cutouts also appear as masks of different sizes within each image. Samples from the synthetic dataset augmented using NoisyCutout, as well as a comparison with images augmented using existing methods: salt-and-pepper noise followed by Cutout, are shown in Fig.~\ref{fig:syn}.

\subsection{Base CNN Architecture}

We choose MobileNetV2 \cite{R43} as our base CNN architecture as it is computationally efficient and can achieve high classification accuracy. We modify its input shape to 299 x 299 x 1 and its classification layer to predict \begin{math}m\end{math} (whether the marker is detected). We also add regression layers that predict 7 other outputs: \begin{math}u\end{math} (horizontal pixel coordinate), \begin{math}v\end{math} (vertical pixel coordinate), \begin{math}d\end{math} (marker pixel diameter) and \begin{math}w, p, q, r\end{math} (quaternions; orientation). While the focus of our work is on localisation, we find that adding an auxiliary task of predicting orientation helps to improve overall performance.

\section{Loss Function}

\subsection{Learning Binary Classification and Pixel Prediction}

We use binary cross-entropy loss for our class predictions as it is commonly used for binary class prediction tasks (we only have two classes to predict). For the regression outputs, we use L2-norm and an object-factor that prevents regression losses from wrongly contributing to the total loss when no marker is present. Our loss function can then be defined as:

\begin{equation}
\mathcal{L}(I) =  BCE(m,\hat{m}) + \varepsilon[\gamma\mathcal{L}_c(I) + \mu\mathcal{L}_d(I) + \mathcal{L}_q(I)]
\label{eq:1}
\end{equation}

where \begin{math}BCE\end{math} represents the binary cross-entropy function for classification loss, \begin{math}\mathcal{L}_c(I) = \|c-\hat{c}\|_2\end{math}, \begin{math}\mathcal{L}_d(I) = \|d-\hat{d}\|_2\end{math} and \begin{math}\mathcal{L}_q(I) = \|q-\hat{q}\|_2\end{math}. \begin{math}\hat{m}\end{math} represents the predicted class, \begin{math}\hat{c}\end{math} the predicted marker pixel coordinates \begin{math}(u,v)\end{math}, \begin{math}\hat{d}\end{math} the predicted marker diameter in pixels and \begin{math}\hat{q}\end{math} the predicted marker orientation vectors in quaternions while \begin{math}m, c, d\end{math} and \begin{math}q\end{math} represent their ground truth. \begin{math}\gamma\end{math} and \begin{math}\mu\end{math} are hyperparameters to balance pixel losses with other loss components and we find \begin{math}\gamma=10\end{math} and \begin{math}\mu=5\end{math} give us the best results in our experiments. \begin{math}\varepsilon\end{math} is the object-factor whose value is 0 when no marker is present in the image and 1 when a marker is present.

\subsection{Leveraging Propagation of Uncertainty}

\begin{figure}[t]
\centerline{\includegraphics[width=0.7\columnwidth]{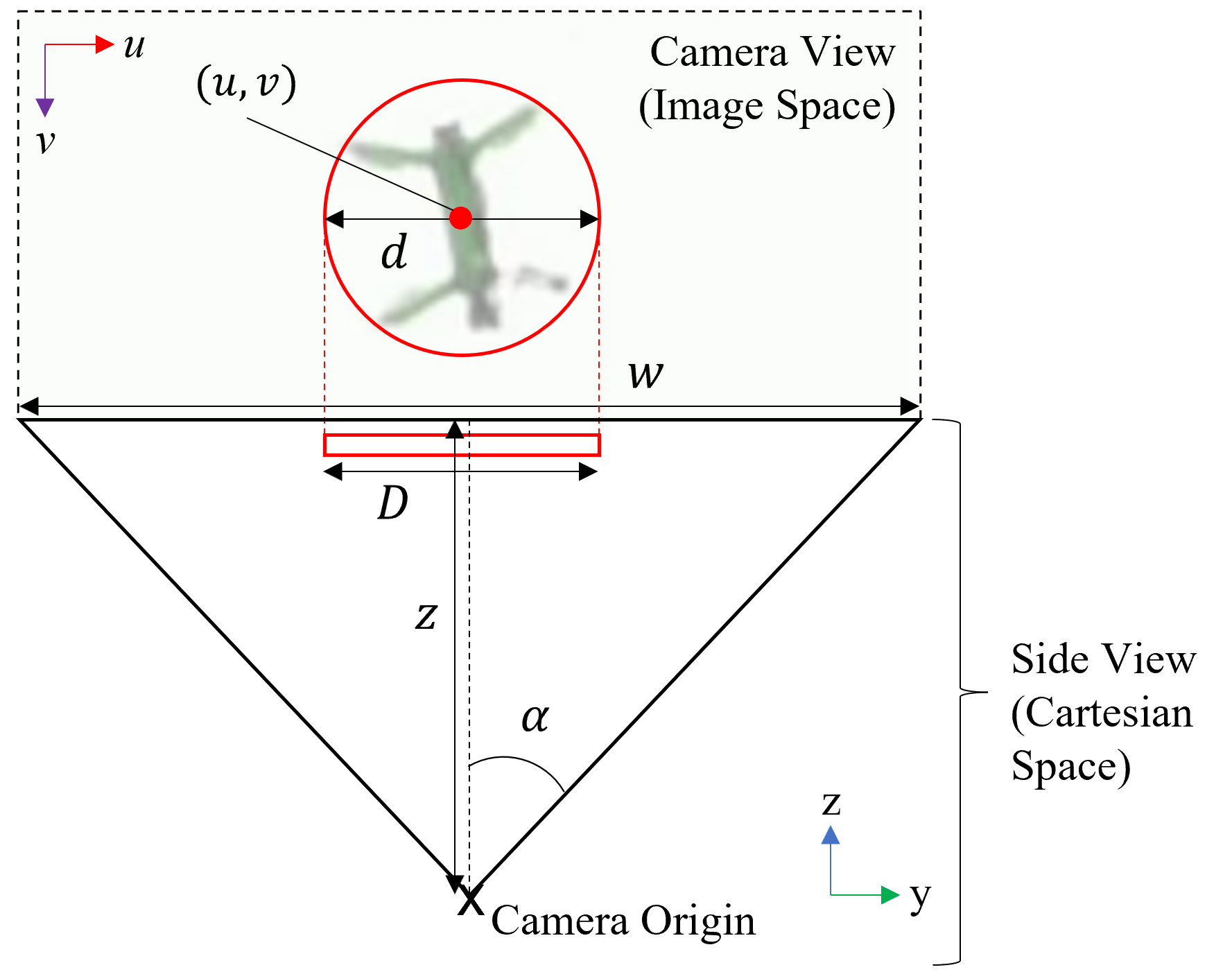}}
\vspace{-1mm}
\caption{Camera view (UAV not to scale) with $u$ and $v$ as horizontal and vertical pixel coordinates within the image, and side view representing the Cartesian space. The observed marker within the camera view is enlarged for better visualisation, with $d$ as the marker's diameter in pixels and $w$ as the total pixels along the image's width. D is the marker’s fixed metric diameter, $\alpha$ is half of the camera’s horizontal FOV (related to $w$), and $z$ is the z-coordinate of the marker from the camera’s origin.}
\label{fig:variables}
\end{figure}

Loss function (1) is formulated with the assumption that the contributions from each unit pixel error from marker pixel coordinates and diameter predictions are equal regardless of the predicted values. However, the relationship between each pixel coordinate in the image space and its coordinate in Cartesian space is non-linear. Intuitively, the higher up a marker is from the camera (larger z value), the smaller it appears in the image space (spanning fewer pixels) and the more each unit pixel contributes to estimation error along the z-axis. With reference to Fig.~\ref{fig:variables}, the z-coordinate of the marker relative to the camera’s origin can be obtained using:

\begin{equation}
z = \frac{Dw}{2d}\cot{\alpha}.
\label{eq:2}
\end{equation}

Based on propagation of uncertainty and equation (2), the relationship between errors in predicted $z$ (\begin{math}\Delta z\end{math}) and small errors in predicted $d$ (\begin{math}\Delta d\end{math}) can be approximated as:

\begin{equation}
\Delta z = \frac{\Delta d}{d}z.
\label{eq:3}
\end{equation}

Combining equations (2) and (3) gives us this relationship:

\begin{equation}
\Delta z \propto \frac{\Delta d}{d^2}.
\label{eq:4}
\end{equation}

This suggests that a decrease in the marker’s actual pixel diameter \begin{math}d\end{math} in the camera’s view quadratically increases the effect \begin{math}\Delta d\end{math} has on \begin{math}\Delta z\end{math}. In a similar manner, using the y-axis as an example, the y-coordinate of the marker relative to the camera’s origin can be obtained using:

\begin{equation}
y = \frac{uD}{2d},
\label{eq:5}
\end{equation}

resulting in this based on propagation of uncertainty:

\begin{equation}
\Delta y = \frac{\Delta u}{u}y.
\label{eq:6}
\end{equation}

Combining equations (5) and (6) gives this relationship:

\begin{equation}
\Delta y \propto \frac{\Delta u}{d}.
\label{eq:7}
\end{equation}

Equations (4) and (7) provide a relationship between pixel errors in the image space and metric error in the Cartesian space using only variables in the image space. We leverage these relationships to improve our loss function by directly relating \begin{math}\mathcal{L}_c(I)\end{math} with the Euclidean error of \begin{math}\Delta u\end{math} and \begin{math}\Delta v\end{math}, and \begin{math}\mathcal{L}_d(I)\end{math} with \begin{math}\Delta d\end{math}, which results in a modified loss function:

\begin{equation}
\mathcal{L}(I) =  BCE(m,\hat{m}) + \varepsilon[\gamma\frac{\mathcal{L}_c(I)}{d} + \mu\frac{\mathcal{L}_d(I)}{d^2} + \mathcal{L}_q(I)]
\label{eq:8}
\end{equation}

We refer to our full CNN-based marker localisation pipeline (including moving object detection) as Marker-based-UAV-Localisation-Net (MbULNet), with the method that is trained using loss function (1) as MbULNet$_{0}$ and the other method that uses loss function (8) leveraging uncertainty propagation (UP) as MbULNet$_{\mathrm{UP}}$.

\section{Experimental Study}

\subsection{CNN Implementation and Training}
We implemented MbULNet's CNN in TensorFlow and trained using a NVIDA RTX Turbo 2080Ti GPU. All images were normalised such their input pixels range from 0 to 1. Of the 6800 generated synthetic binary images, 5400 were used for training and the remaining 1400 to obtain validation loss during the training process. We trained our CNNs using randomly initialised weights and optimised them using ADAM \cite{R44} with a batch size of 32. For all the CNNs, we experimented with training at different learn rates of between \begin{math}10^{-6}\end{math} to \begin{math}10^{-3}\end{math} and for 150 to 250 epochs, and used the best performance obtained by each CNN for comparison.

\subsection{Experimental Setup}
\begin{figure}[t]
\centerline{\includegraphics[width=0.9\columnwidth]{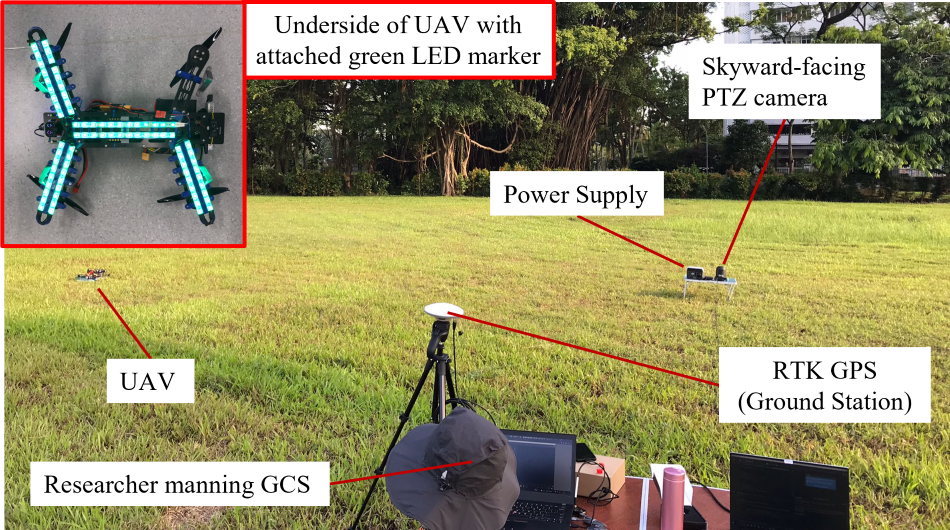}}
\vspace{-1mm}
\caption{Experimental setup consisting of a UAV attached with a green LED marker, a Ground Control Station (GCS) to set waypoints, and a Panasonic AW-HE40H PTZ camera tilted 90{\degree} skywards to observe the UAV. RTK GPS was used to obtain ground truth and is not part of our methodology.}
\label{fig:setup}
\end{figure}

We designed an LED marker with a diameter $D$ of 0.47m such that it maximises the footprint of the UAV used in our experiments (less an arm so that it is non-symmetrical for distinguishing orientation). We deployed a Panasonic AW-HE40H PTZ camera tilted skywards at 90{\degree} from the ground and set its brightness to its lowest level as we find this helps the LED marker appear clearer in the camera. A GCS was used to set waypoints for the UAV while RTK GPS was used to obtain ground truth at about 5 Hz and is not part of our methodology. Fig.~\ref{fig:setup} shows our experimental setup.

We collected data from four real-world outdoor flights in the daytime where the UAV flew an approximately 4m x 4m square profile via waypoints. A different combination of height and camera zoom (horizontal FOV) was used for each flight and are stated in Table~\ref{tab1}. We evaluated three methods: 1) MbULNet$_{0}$(SPCutout) trained on synthetic binary images augmented using existing salt-and-pepper and Cutout; 2) MbULNet$_{0}$(NoisyCutout) trained on synthetic binary images augmented using our proposed NoisyCutout; and 3) MbULNet$_{\mathrm{UP}}$(NoisyCutout) trained on the same dataset used for MbULNet$_{0}$ but with a loss function leveraging uncertainty propagation, by testing them on videos recorded at 12.5 FPS (limited by the camera at HD resolution) to predict the UAV's Cartesian coordinates for each flight.

We tested different values of threshold, $\tau$, described in section \ref{locpipeline} to find the value that resulted in the best performance for each method when kept constant across all flights (we used $\tau = 25$ for MbULNet$_{0}$(SPCutout), $\tau = 10$ for MbULNet$_{0}$, and $\tau = 13$ for MbULNet$_{\mathrm{UP}}$). On a laptop with an NVIDIA GeForce GTX 1650 Ti GPU, the CNN in MbULNet had an average inference time of about 30ms while the full pipeline ran at about 20 Hz. All predicted z-coordinates were smoothed using a 1st order Butterworth filter (critical frequency = 1Hz) and did not substantially affect our evaluation. We interpolated all localisation predictions at time intervals of 0.01s and used values closest to timestamps of the ground truths for evaluation. A total of 169 ground truth points were extracted across all flights for evaluation. Three experiments were conducted: 1) to evaluate whether using the proposed NoisyCutout to augment binary images for training MbULNet$_{0}$ improves localisation performance; 2) to evaluate whether leveraging uncertainty propagation in the loss function of MbULNet$_{\mathrm{UP}}$ improves localisation performance as compared to MbULNet$_{0}$; and 3) to evaluate the robustness of marker detection against various backgrounds and the possibility of closed-loop control.

\begin{figure}[t]
\centerline{\includegraphics[width=0.9\columnwidth]{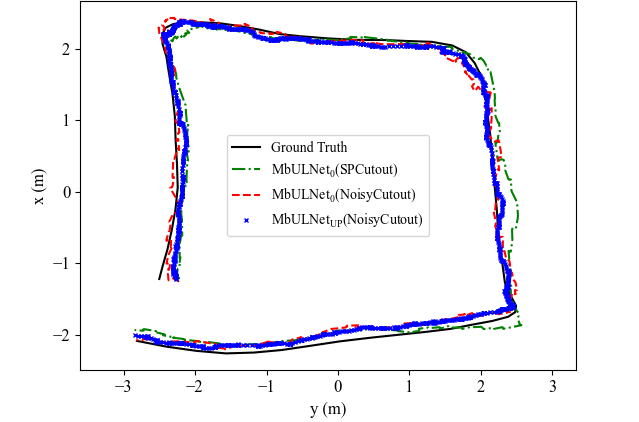}}
\vspace{-3mm}
\caption{Top view (x-y plane) localisation results for one of the UAV flights (at about 15m height from the camera and at horizontal FOV of 54°). All estimated flight paths are close to the ground truth, showing that we can successfully detect and localise a UAV using our proposed MbULNet.}
\label{fig:xyplot}
\vspace{-1mm}
\end{figure}

\begin{figure}[t]
\centerline{\includegraphics[width=0.9\columnwidth]{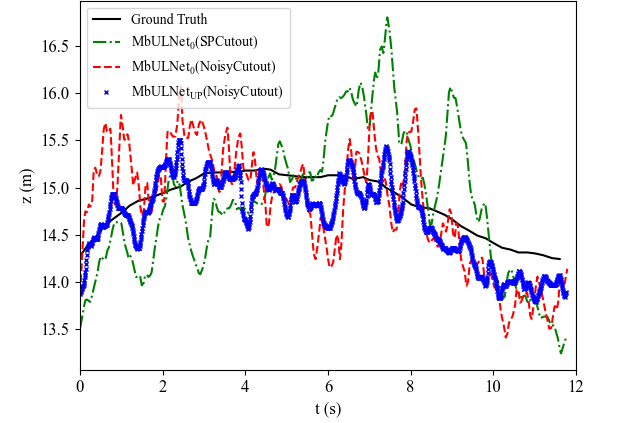}}
\vspace{-3mm}
\caption{z-coordinate localisation results comparing all methods for the same UAV flight in Fig.~\ref{fig:xyplot}. Training our CNN on images augmented with the proposed NoisyCutout (red) results in smaller z-fluctuations about the ground truth as compared to using existing methods (green). Our proposed MbULNet$_{\mathrm{UP}}$ (blue) obtains z-estimates closest to the ground truth.}
\label{fig:zplot}
\end{figure}

\section{Results and Discussion}

\subsection{Marker-based UAV Localisation with Deep Learning}

\textit{1) CNN-based UAV Localisation Without Training on Real Images:} We successfully localise the UAV in all four flights with all three methods used in our evaluation. Fig.~\ref{fig:xyplot} shows the flight path and x-y localisation results for one of our flights (Flight D), while Fig.~\ref{fig:zplot} shows results along the z axis. Qualitative results for all three methods show that our proposed CNN-based pipeline successfully detects the UAV, tracks its flight profile and predicts z-coordinates similar to the ground truth for all flights. This is observed across all flights and suggests that the synthetic binary images used for training the CNN successfully simulates binary images processed from real images for our task. We can conclude that our CNN-based method successfully achieves real-world marker-based UAV localisation without the need to train on any real images and demonstrate sim-to-real transfer for detection of the LED marker and size prediction.

\textit{2) Localisation Accuracy:} Quantitative results in Table~\ref{tab1} show that we achieve an overall 3D root-mean-square-error (RMSE) of 0.407m using the best-performing combination of our methods, MbULNet$_{\mathrm{UP}}$(NoisyCutout). This is a substantial improvement compared to results reported in other recent works using CNN-based UAV localisation \cite{R30,R31,R32,R37} with reported RMSE ranging between 0.65m to 4m. Our results also show that the 3D RMSE for each flight is very close (about 0.03m) to their respective z-coordinate RMSE compared to their 2D RMSE, suggesting that errors in z-coordinate contribute largest to the overall localisation error. 


\captionsetup[table]{position=bottom}

\newcommand\boldblue[1]{\textcolor{blue}{\textbf{#1}}}

\newcolumntype{C}{>{\centering\arraybackslash}X}
\newcolumntype{R}{>{\raggedright\arraybackslash}X}

\begin{table}[t]
\vspace{+2mm}
\begin{center}
\begin{tabularx}{1.0\columnwidth}{m{1.5cm} m{0.7cm} m{1.5cm} m{1.5cm} m{1.5cm}}
\hline
 \centering Flight & \centering RMSE
 & \centering MbULNet$_{\mathrm{0}}$ SPCutout (Existing) & \centering MbULNet$_{\mathrm{0}}$ NoisyCutout (Proposed) & \centering\arraybackslash MbULNet$_{\mathrm{UP}}$ NoisyCutout (Proposed) \\
\hline
 \multirow{3}{1.5cm}{\centering A) 13.87 m HFOV: 46{\degree}} & \centering 2D & \centering 0.237 m & \centering 0.181 m & \centering\arraybackslash \boldblue{0.167 m} \\ 
 & \centering z & \centering 0.886 m & \centering 0.454 m & \centering\arraybackslash \boldblue{0.416 m} \\ 
 & \centering 3D & \centering 0.917 m & \centering 0.489 m & \centering\arraybackslash \boldblue{0.449 m} \\ 
\hline
 \multirow{3}{1.5cm}{\centering B) 18.92 m, HFOV: 35.9{\degree}} &
 \centering 2D & \centering 0.255 m & \centering 0.139 m & \centering\arraybackslash \boldblue{0.125 m} \\ 
 & \centering z & \centering 1.311 m & \centering 0.650 m & \centering\arraybackslash \boldblue{0.454 m} \\ 
 & \centering 3D & \centering 1.335 m & \centering 0.665 m & \centering\arraybackslash \boldblue{0.471 m} \\ 
\hline
 \multirow{3}{1.5cm}{\centering C) 9.14 m, HFOV: 54{\degree}} &
 \centering 2D & \centering 0.238 m & \centering 0.166 m & \centering\arraybackslash \boldblue{0.157 m} \\ 
 & \centering z & \centering 0.679 m & \centering 0.421 m & \centering\arraybackslash \boldblue{0.404 m} \\ 
 & \centering 3D & \centering 0.719 m & \centering 0.453 m & \centering\arraybackslash \boldblue{0.433 m} \\ 
\hline
 \multirow{3}{1.5cm}{\centering D) 14.83 m HFOV: 54{\degree}} &
 \centering 2D & \centering 0.150 m & \centering 0.177 m & \centering\arraybackslash \boldblue{0.140 m} \\ 
 & \centering z & \centering 0.630 m & \centering 0.460 m & \centering\arraybackslash \boldblue{0.274 m} \\ 
 & \centering 3D & \centering 0.647 m & \centering 0.493 m & \centering\arraybackslash \boldblue{0.308 m} \\ 
\hline
 \multirow{3}{1.5cm}{\centering Overall} &
 \centering 2D & \centering 0.216 m & \centering 0.168 m & \centering\arraybackslash \boldblue{0.147 m} \\ 
 & \centering z & \centering 0.893 m & \centering 0.502 m & \centering\arraybackslash \boldblue{0.380 m} \\ 
 & \centering 3D & \centering 0.919 m & \centering 0.530 m & \centering\arraybackslash \boldblue{0.407 m} \\ 
\hline

\end{tabularx}
\end{center}
\vspace{-3mm}
\caption{Localisation root-mean-square-error (RMSE) at different mean UAV height-from-camera and horizontal FOV (HFOV). We compare MbULNet$_{0}$ trained on images augmented using existing methods to using our proposed NoisyCutout, and MbULNet$_{0}$ to MbULNet$_{\mathrm{UP}}$.}
\vspace{-1mm}
\label{tab1}
\end{table}

\textit{3) Deployability:} Our method works with varying values of known camera resolution and FOV, suggesting that this method may be used with any camera of known intrinsic parameters that is set up in the same manner. This also suggests that our method has the potential to localise UAVs operating at higher altitudes (beyond 20m) by increasing the zoom level of the camera (reducing HFOV) or using cameras of higher resolution. This flexibility, with the use of synthetic data instead of real data for training, makes our method highly deployable for real-world tasks.

\subsection{Comparison between NoisyCutout and Existing Methods}

\begin{figure}[t]
\centerline{\includegraphics[width=\columnwidth]{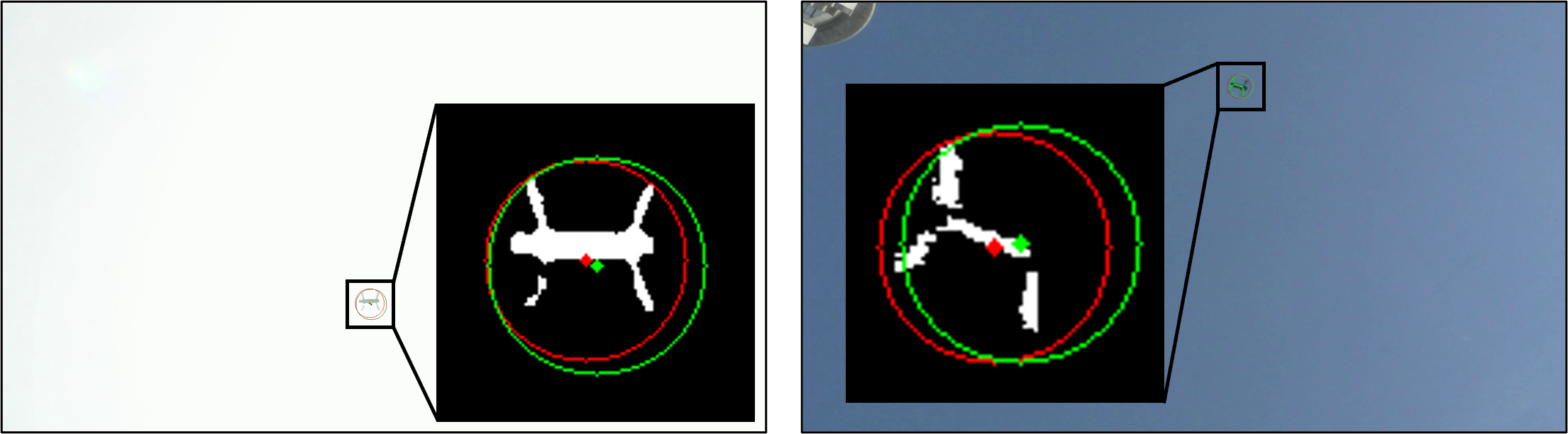}}
\vspace{-1mm}
\caption{Example predictions extracted from different flights with their enlarged binary images after thresholding for input into the CNN. We observe that in both examples, MbULNet$_{0}$(SPCutout) (green circles) overestimates the marker’s size while MbULNet$_{0}$(NoisyCutout) (red circles) predicts better marker size and centroid position.}
\label{fig:spvsnoisycutout}
\vspace{-1mm}
\end{figure}

The results in Table~\ref{tab1} show that MbULNet$_{0}$(NoisyCutout) achieves lower z-coordinate and 3D RMSE across all flights as compared to MbULNet$_{0}$(SPCutout). Fig.~\ref{fig:spvsnoisycutout} shows examples of marker location and size estimation by these two methods and we observe that MbULNet$_{0}$(SPCutout) overestimates the size of the marker while MbULNet$_{0}$(NoisyCutout) estimates better marker size and centroid position. We conclude that augmenting binary images using the proposed NoisyCutout improves the CNN’s ability to learn to locate and size the marker from thresholded real images as compared to using existing salt-and-pepper and Cutout methods.

\begin{figure}[t]
\centerline{\includegraphics[width=\columnwidth]{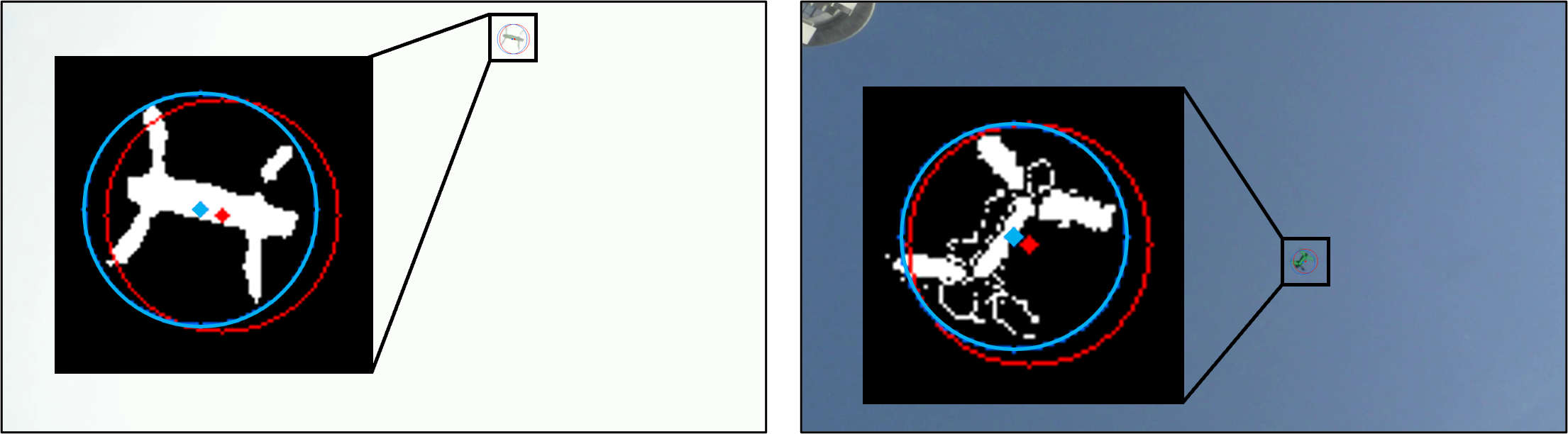}}
\vspace{-1mm}
\caption{Example predictions extracted from different flights with their enlarged binary images after thresholding for input into the CNN. We observe that in both examples, MbULNet$_{\mathrm{UP}}$(NoisyCutout) (blue circles) predicts a more accurate marker boundary and centroid as compared to MbULNet$_{0}$(NoisyCutout) (red circles).}
\label{fig:0vsup}
\vspace{-1mm}
\end{figure}

\subsection{Loss Function with Propagation of Uncertainty}

We compare MbULNet$_{0}$ and MbULNet$_{\mathrm{UP}}$ both trained on the same images augmented using NoisyCutout. We observe in Fig.~\ref{fig:zplot} that MbULNet$_{\mathrm{UP}}$ results in better z-coordinate estimates as they fluctuate with a smaller amplitude and are generally closer to the ground truth compared to estimates by MbULNet$_{0}$. Fig.~\ref{fig:0vsup} shows example predictions by MbULNet$_{\mathrm{UP}}$ (blue circles) and MbULNet$_{0}$ (red circles) and show that MbULNet$_{\mathrm{UP}}$ predicts more accurate marker diameters and centroid positions. Table~\ref{tab1} also shows that MbULNet$_{\mathrm{UP}}$ reduces all localisation RMSE across all flights. Overall, MbULNet$_{\mathrm{UP}}$ reduced 2D RMSE by 12.5\%, z-coordinate RMSE by 24.3\% and 3D RMSE by 23.2\% as compared to using MbULNet$_{0}$. We conclude that using a loss function that leverages uncertainty propagation relating pixel error to metric error can help a CNN to learn this relationship and improve predictions of the marker’s position and observed diameter, achieving better 3D localisation accuracy while under real-world effects described in this work.

\begin{figure}[t]
\centerline{\includegraphics[width=\columnwidth]{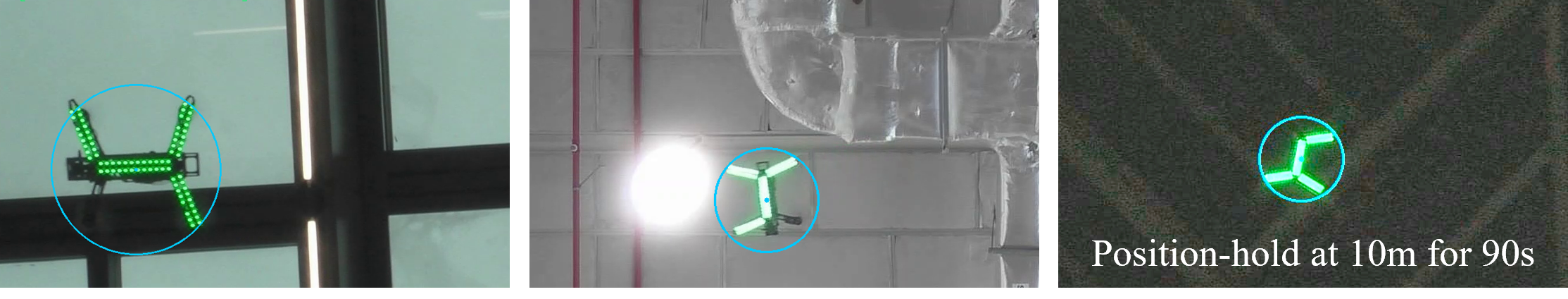}}
\vspace{-1mm}
\caption{Example detections of the UAV with LED marker against various challenging backgrounds within indoor environment. \textit{Left:} Room with glass ceiling. \textit{Middle:} Fabrication facility. \textit{Right:} Exhibition hall - where we also demonstrate a 90s hover using our method as the only source of external positioning system to show that closed-loop control is possible.}
\label{fig:robust_hover}
\end{figure}

\subsection{Detection Robustness and Closed-Loop Control}
We demonstrate tracking of the UAV against different backgrounds and Fig.~\ref{fig:robust_hover} shows examples of successful detection using MbULNet$_{\mathrm{UP}}$, demonstrating the robustness of our method even with substantial background clutter and light sources. On the other hand, the performance of our method against a background with moving features (e.g. trees) and tracking of multiple UAVs concurrently have yet to be explored. We also demonstrate that a UAV can perform a 90 seconds hover at 10m from the ground using our method as the only external positioning system to show that closed-loop control is possible.

\section{Conclusion}
In this work, we demonstrate marker-based UAV localisation with our proposed CNN-based localisation pipeline that uses only a single ground-based camera. Our method is highly deployable for real-world applications as it only requires synthetic images for training the CNN and can be easily adapted for use with any camera given its intrinsic parameters. We propose NoisyCutout algorithm to augment synthetic binary images to simulate such images processed from real images and show that it is more effective than using existing augmentation methods. We show that a CNN can learn a relationship between image space and Cartesian space by leveraging the concept of uncertainty propagation in its loss function and demonstrate that this improves localisation performance through real-world experiments. We also demonstrate its robustness in different backgrounds and that closed-loop control is possible with our method.

Future work will explore how to optimise the choice of parameters such as FOV, resolution, and marker size at varying UAV-to-camera distances, as well as expand the operational range of the system using an active PTZ camera.

\bibliographystyle{./bibliography/IEEEtran}
\bibliography{./bibliography/main}

\end{document}